# PLANET v2.0: A comprehensive Protein-Ligand Affinity Prediction Model Based on Mixture Density Network

Haotian Gao, [#] Xiangying Zhang, [#] Jingyuan Li, Xinchong Chen, Haojie Wang, Yifei Qi,* and Renxiao Wang*

*Department of Medicinal Chemistry, School of Pharmacy, Fudan University, 826 Zhangheng Road, Shanghai 201203, People's Republic of China*

[#] These authors had equal contributions to this work.

* To whom all correspondence should be addressed: *yfqi@fudan.edu.cn* (Y. Qi); *wangrx@fudan.edu.cn* (R. Wang)

## Abstract

Drug discovery represents a time-consuming and financially intensive process, and virtual screening can accelerate it. Scoring functions, as one of the tools guiding virtual screening, have their precision closely tied to screening efficiency. In our previous study, we developed a graph neural network model called PLANET (Protein-Ligand Affinity prediction NETwork), but it suffers from the defect in representing protein-ligand contact maps. Incorrect binding modes inevitably lead to poor affinity predictions, so accurate prediction of the protein-ligand contact map is desired to improve PLANET. In this study, we have proposed PLANET v2.0 as an upgraded version. The model is trained via multi-objective training strategy and incorporates the Mixture Density Network to predict binding modes. Except for the probability density distributions of non-covalent interactions, we innovatively employ another Gaussian mixture model to describe the relationship between distance and energy of each interaction pair and predict protein-ligand affinity like calculating the mathematical expectation. As on the CASF-2016 benchmark, PLANET v2.0 demonstrates excellent scoring power, ranking power, and docking power. The screening power of PLANET v2.0 gets notably improved compared to PLANET and Glide SP and it demonstrates robust validation on a commercial ultra-large-scale dataset. Given its efficiency and accuracy, PLANET v2.0 can hopefully become one of the practical tools for virtual screening workflows. PLANET v2.0 is freely available at https://www.pdbbind-plus.org.cn/planetv2.

## 1. Introduction

As the rapid growth of make-on-demand chemical libraries[1-4], typical experimental approaches[5, 6] to identifying actives of specific targets are difficult to implement[7]. From another perspective, large-scale libraries provide opportunities to thoroughly explore chemical space through virtual screening[8], which is more efficient and relatively low-cost. Virtual screening can be categorized into ligand-based virtual screening (LBVS) and structure-based virtual screening (SBVS)[9, 10]. Ligand-based approaches output molecules with similarity to known active compounds[11, 12], and the advantage of library scale cannot be fully exploited[13]. SBVS is a process that evaluates binding affinities to a specific target of candidate molecules through molecular docking and ranks these molecules correctly under the guidance of protein-ligand affinity prediction[14, 15]. With the help of Alphafold[16], obtaining protein structures is no longer a problematic issue for structure-based methods[17-19]. Therefore, prediction of binding free energy becomes the critical factor for the success rate of virtual screening. To improve the efficiency of virtual screening, it is imperative to achieve a balance between prediction precision and computational cost.

Scoring functions calculate binding affinity efficiently through some empirical approximations and are suitable for high-throughput screening. They are typically classified into four categories[20]: physics-based, knowledge-based, empirical, and machine learning functions (MLSFs). Physics-based methods utilize physical energy terms, whose training data is always from experiments and quantum-mechanical calculations. Knowledge-based methods estimate the binding energy based on statistical results from large structure-affinity databases. Empirical methods are widely used today, such as Glide SP[21], Autodock Vina[22] and X-Score[23], which describe protein-ligand interactions as the summation of weighted interaction energy terms. Although these methods have proven useful in the realm of drug discovery, they still have certain limitations. As pointed out before, conventional scoring functions suffer from a high false positive rate in their results[24]. Also, the traditional formulas are too simple, predominantly employing linear combinations, to represent the complicated relationship between protein-ligand complexes and binding affinities[25].

Machine learning scoring functions, especially deep learning scoring functions (DLSFs), are essentially a high-dimensional representation of structure-affinity pairs[26-28]. They do not have predefined formulas so that they can fit the training data more flexibly and always demonstrate superior predictive accuracy, emerging as the mainstream in this field[29-34]. Comparative analysis between conventional scoring functions tested in CASF-2016[24] and recently reported studies reveals that these novel approaches generally demonstrate a 10% improvement in scoring power[35-37]. However, a significant characteristic of these methods is their tendency to fit the distribution of training data, so the quality and quantity of training data determine the upper bound of their performance. In the field of scoring functions, the available data remain relatively small in scale with a substantial disparity compared to other fields. It is universally acknowledged that measurements of binding affinity are sensitive to experimental conditions, so errors inevitably exist in datasets[38, 39], resulting in constraints for the development of scoring functions. Another issue is the preference of tasks[40], meaning that some methods produce precise predictions on one task, but poorly perform in other application scenes.

A perfect scoring function should be generalized and obtain good results across all tests in CASF-2016, including scoring power, ranking power, docking power, and screening power. To turn this goal into reality, data augment is one of the typical approaches, where different types of negative samples are added into training set[41-46]. Besides, some models' output head is split and each head is trained for a specific task[47-49]. In our opinion, multiple heads will make users confused because scoring function's application scenarios are usually composite. Optimizing the representation of interactions is another strategy, and Mixture Density Network (MDN)[50] introduced by DeepDock[51] is the most impressive one among studies of this methodology. DeepDock decomposes protein-ligand interactions into sets of surface vertex-atom interaction pairs and uses MDN to learn the probability density distribution of each pair distance. Then, these distributions are aggregated into a statistical potential to score a protein-ligand complex. RTMScore proposed by Shen *at el*.[52] follows a similar approach and advances MDN from the perspective of coarse-grained protein representation. However, these two models are only good at docking power and screening power and need further advances. As an updated version of RTMScore, GenScore[53] achieves improvements in scoring and ranking by add the correlation between statistical potentials and affinities into its loss function. Thus, the values predicted by GenScore only allow for relative comparisons and do not represent actual binding affinities. Other MDN-based models, such as Interformer[54] and DiffDock-NMDN[55], obtain the ability to score by introducing additional scoring modules. Recently, a novel model named RefineScore[40] integrates distance likelihood with empirical scoring functions to address all four tasks, remaining within the MDN framework.

Therefore, no comprehensive end-to-end protein-ligand affinity prediction model has been reported to date. In this study, we introduce PLANET v2.0, a novel MDN-based protein-ligand affinity prediction model. It addresses several unresolved issues in PLANET[41], which is called PLANET v1.0 in this paper. Unbalanced binary labels of interactions are replaced by probability density distributions, so that PLANET v2.0 achieves a significant improvement in docking power. Beyond modeling probability density distribution of pairwise distances, we employ modified Gaussian mixture models (GMMs) to fit the relationship between interaction distances and binding energies, where the mixture coefficients can assume positive or negative values to represent favorable and unfavorable interactions respectively. The parameters required for the energy function are consistent with those of the distance density function, so they can be integrated into the MDN framework. PLANET v2.0 inherits the input format and multi-objective training strategy of our last model, so our model's generalizability and interpretability get guaranteed. PLANET v2.0 was evaluated on the CASF-2016 benchmark in terms of scoring power, ranking power, and docking power. LIT-PCBA[56] was used to test our model's performance in virtual screening. In order to further assess screening power in the real drug discovery scene, an ultra-large-scale virtual screening dataset was constructed with LIT-PCBA and a commercial compound library, thereby enhancing the challenge level of the evaluation. Our performance in lead optimization was also tested in this study. Across all benchmarks, PLANET v2.0 achieves state-of-the-art performance or results comparable to conventional scoring functions, demonstrating its versatility as a comprehensive scoring function.

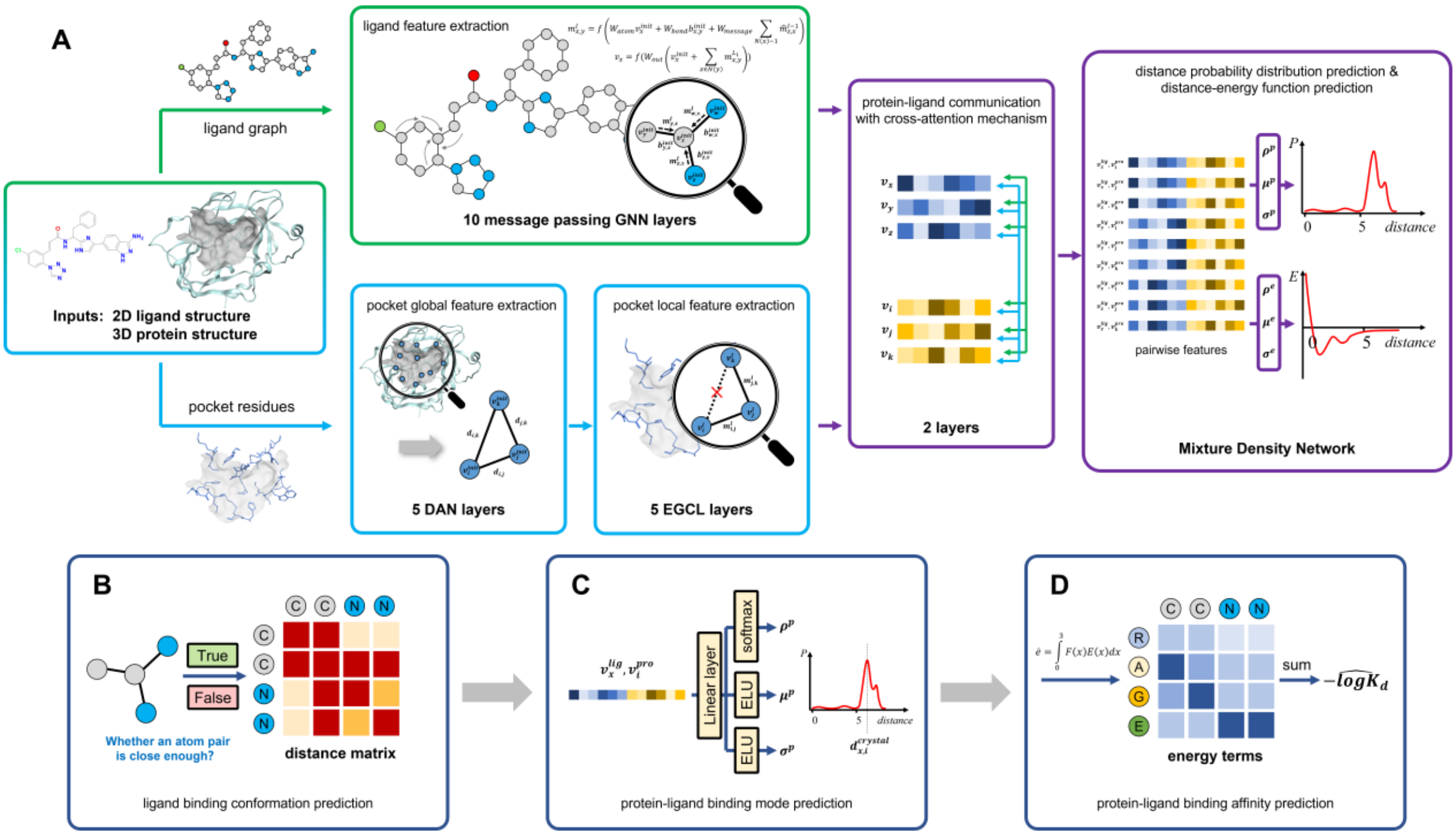

**Figure 1**. Illustration of the basic framework of PLANET v2.0. (A) Overall architecture of PLANET v2.0. Proteins and ligands are represented as graphs, then go through different graph networks to learn their node representations. After updating features with the cross-attention mechanism, node features are concatenated in a pairwise manner and then processed with Mixture Density Network to predict efficient of Gaussian mixture models. Two types of Gaussian models are constructed to model the relationship between distance, probability, binding energy. (B-D) Three training tasks of multi-objective learning strategy: derivation of the ligand distance matrix(B), prediction of atom-residue distance probability density distribution (C), and prediction of protein-ligand binding

affinity (D).

## 2. Results

### 2.1 Prediction of protein-ligand binding affinity

One of the primary applications of the PLANET v2.0 is to predict binding affinities between protein-ligand complexes accurately. The scoring performance of PLANET v2.0 is tested on CASF-2016, a widely recognized benchmark in the field of scoring functions. The correlation between experimental affinities (-log$K_d$/$K_i$) and predicted affinities is shown in Figure 2, where the mean absolute error is 0.910, the root mean square error (RMSE) is 1.171 and the Pearson correlation coefficient (Pearson $R$) is 0.848. PLANET v2.0 significantly surpasses the first-generation model (Pearson $R$ is 0.824), and lead among all machine learning/deep learning models and physics-based scoring functions, the performances of other methods are summarized in Table 1. In addition to MDN-based methods, the others in Table 1 were selected based on their publish year and architectures. These ML/DL methods mostly achieve Pearson $R$ within the range of 0.80 to 0.85 and the performances of conventional functions are mostly lower than 0.8. It is shown that our model produces competitive performance of protein-ligand binding affinity predicting based on the current scale of data.

It is worth noting that PLANET v2.0 was trained and tested in the strictest way, so that the rank of our model among all methods may be underestimated. Firstly, similar data in the training set and CASF-2016 was removed, while the other methods do not implement such steps, leading their performance to be overestimated. The procedures used here can be seen in *Methods*. Secondly, our model does not take three-dimensional protein-ligand structures as input, which means fewer features can be learned directly. As we pointed before, the scoring power of those methods in Table 1 may decrease considerably if complex structures derived from molecular docking are used. All features of protein-ligand interactions are extracted and modelled inside PLANET v2.0, making our model not rely on the quality of complex structures. Thus, it is still necessary to develop models with strong generalization and robustness like PLANET v1.0 and v2.0, although it is indicated that 3D binding pose is essential to affinity prediction[57]. Last but not least, the affinity predicted by PLANET v2.0 is in the form of summation of energy terms, instead of a value predicted from a fully-connected network. This function formula is similar with conventional scoring functions, so that our model can produce predicted values within a boarder range, while DL models tend to provide predicted values that conform to the distribution of the training set labels. Certain models in Table 1 may exhibit performance overestimation due to affinity value homology between training and test sets. This is an advantage that is difficult to reflect in metrics.

'Ranking power' test is another test related to the accuracy of binding affinity prediction in CASF-2016, and it is more close to the real drug development scenarios. PLANET v2.0 yielded an average Spearman correlation coefficient of 0.669 with the 90% confidence interval of 0.579 to 0.732 on the 57 target proteins in the CASF-2016 benchmark. As shown in Table 2 and our evaluation in CASF-2016[24], Spearman correlation coefficient (Spearman $R$) of PLANET v2.0 is better than all other conventional scoring functions and is comparable to MDN-based methods. Both versions of PLANET demonstrate comparable performance in this benchmark test[41]. It is posited that difference of binding modes and particularly hotspots between protein and ligand constitute the primary determinant of binding capability for different ligands. Although PLANET v2.0 attempted to predict these hotspots through binding energy distribution modeling, the approach remains inferior to directly utilizing complex structures as input. As one can see, most of MDN-based methods were not good at affinity prediction, Interformer employs an additional deep learning network to predict binding affinity based on MDN-derived complex structures, which enhances its scoring performance but compromises robustness when handling structures of varying quality. Without requiring integration with other models, the innovative architecture of PLANET v2.0 partially addressed this issue.

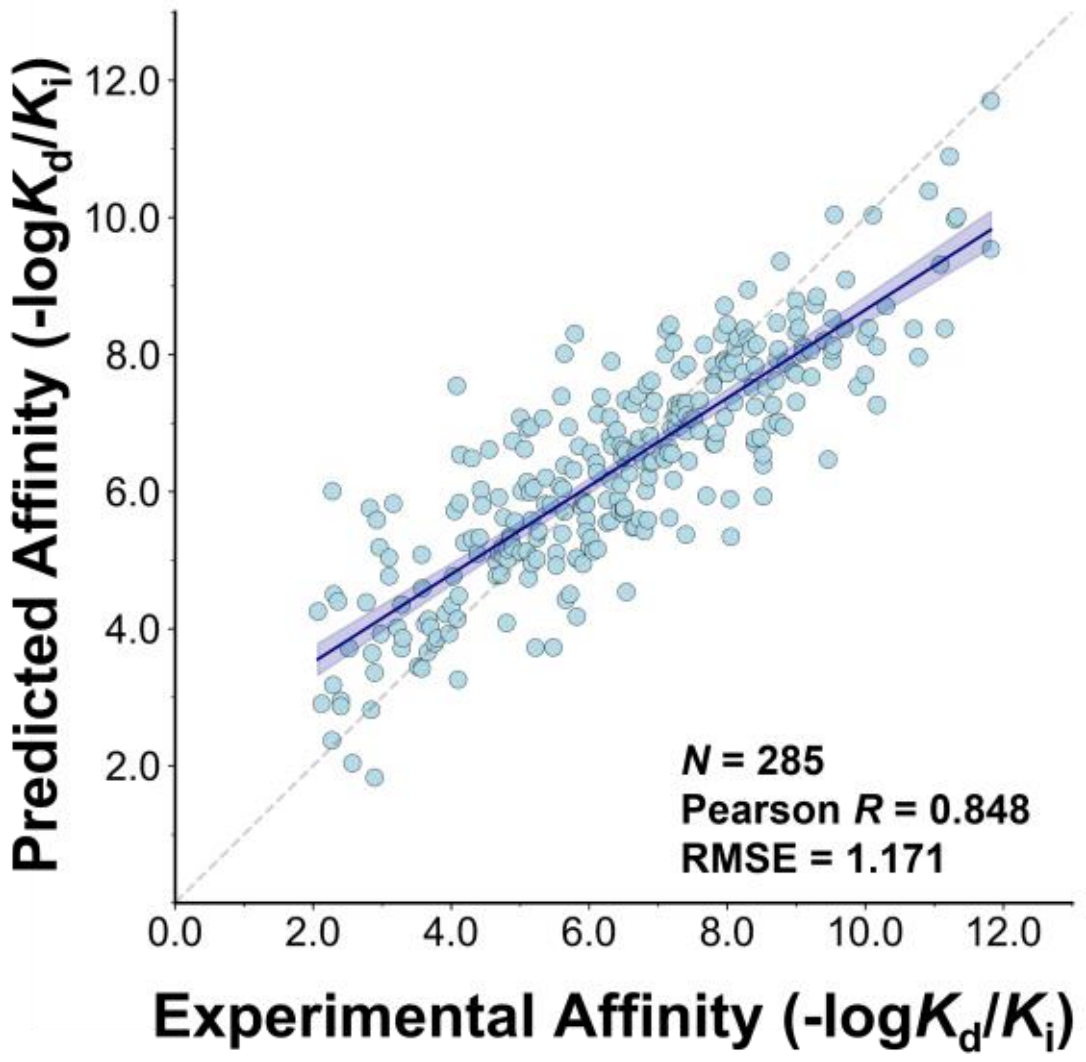

**Figure 2.** Correlation between the experimental and the predicted values by PLANET v2.0 on the CASF-2016 test set. The solid line is the regression line, where the 95% confidence interval is indicated by the shaded region.

**Table 1.** Scoring power of several scoring functions tested on the CASF benchmark.

| Publish Year | Model | Task-specificity | Training Set | RMSE | Pearson *R* |
|---|---|---|---|---|---|
| 2025 | PLANET v2.0[a] | - | PDBbind general set v.2021 (N = 18455) | 1.171 | 0.848 |
| 2023 | PLANET v1.0[41] | - | PDBbind general set v.2020 (N = 15616) | 1.247 | 0.824 |
| 2025 | RefineScore[a 40] | - | PDBbind general set v.2020 (N = 19158) | 1.15 | 0.848 |
| 2024 | DiffDock-NMDN[a 51] | - | PDBbind general set v.2020 (N = 12554) | N.A. | 0.866[b] |
| 2024 | Interformer[a 52] | scoring, ranking, d-ocking | PDBbind general set v.2020 (N = 15686) | N.A. | 0.802 |
| 2023 | GenScore[a 53] | - | PDBbind general set v.2020 (N = 17649) | N.A. | 0.773 |
| 2022 | RTMScore[a 54] | docking, screening | PDBbind general set v.2020 (N = 17649) | N.A. | 0.507 |
| 2021 | DeepDock[a 55] | docking, screening | PDBbind general set v.2019 (N = 15000) | N.A. | 0.525 |
| 2025 | CL-GNN[58] | scoring | BioLip[59] (N = 351458) | 1.200 | 0.838 |
| 2024 | DeepRLI[47] | - | PDBbind general set v.2020 (N = 7337) | 1.176 | 0.849 |
| 2024 | CPIScore[60] | scoring | PDBbind general set v.2019 (N = 12101) | 1.57[c] | 0.74[c] |
| 2024 | ConBAP[61] | - | Composite dataset from PDBbind and redocked (N = 63885) | 1.127 | 0.864 |
| 2023 | GIGN[62] | scoring | PDBbind general set v.2019 (N = 11904) | 1.190 | 0.840 |
| 2022 | Sfcnn[63] | scoring | PDBbind refined set v.2019 (N = 4100) | 1.326 | 0.793 |
| 2022 | DCML[64] | - | PDBbind refined set v.2016 (N = 3772) | 1.255 | 0.843 |
| 2022 | MPNN[65] | scoring | PDBbind general set v.2016 (N = 9662) | 1.511[e] | 0.813[e] |
| 2021 | IGN[66] | - | PDBbind general set v.2016 (N = 8298) | 1.22[d] | 0.837[d] |
| 2021 | 'Late fusion model'[67] | scoring, screening | PDBbind general set v.2016 [f] | 1.326[a] | 0.808[a] |
| 2021 | PerSpect-GBT[68] | scoring | PDBbind refined set v.2016 (N = 3772) | 1.244 | 0.84 |
| 2018 | $K_{DEEP}$[69] | scoring | PDBbind refined set v.2016 (N = 3767) | 1.27 | 0.82 |
| 2018 | Pafnucy[70] | - | PDBbind general set v.2016 (N = 11906) | 1.42[e] | 0.78[e] |
| 2010 | Autodock Vina[22] | - | - | 1.73 | 0.604 |
| 2004 | Glide SP[21] | - | - | 1.89 | 0.513 |
| 2002 | X-Score[23] | - | - | 1.69. | 0.631 |

*a*: These models are mixture density network-based models.

*b*: It was reported with p$K_d$ score as its scoring function, instead of scores based on predicted distributions.

*c*: It was reported that 152 complexes in the CASF-2013 benchmark were considered in this evaluation.

*d*: It was reported that 262 complexes in the CASF-2016 test set were considered in this evaluation.

*e*: It was reported that 290 complexes, including 284 in the standard CASF-2016 test set, were considered in this evaluation.

*f*: The exact number of samples included in the training set was not mentioned in the reference.

**Table 2.** Ranking power of several conventional scoring functions and MDN-based models on the CASF-2016 benchmark.

| Type | Method | Spearman *R* |
|---|---|---|
| Conventional scoring functions | Glide SP | 0.419 |
| | Autodock Vina | 0.528 |
| | X-Score | 0.605 |
| MDN-based deep learning models | DeepDock | 0.495 |
| | RTMScore | 0.563 |
| | GenScore | 0.659 |
| | Interformer | 0.809 |
| | DiffDock-NMDN | 0.458 |
| | RefineScore | 0.723 |
| | PLANET v2.0 | 0.669 |

**2.2 Performance of virtual screening on LIT-PCBA and a modified dataset**

Screening power denotes the scoring function's ability to distinguish binders that have strong affinity for a specific target from non-binders in a chemical library. As we revealed before[41], data in PDBbind dataset tend to concentrate at some certain regions in the space of protein-ligand complexes, while data in DUD-E or LIT-PCBA distribute in a broader area. Thus, it is challenging for machine learning/deep learning methods, especially structure-based methods, to show their potentials in virtual screening because of limited training data. In addition, DUD-E has been noted to suffer potential biases due to the high similarity among molecules in each target. To ensure the objectivity of screening power test, LIT-PCBA was finally chosen as our benchmark, as LIT-PCBA had undergone unbiased processing and validation. Notably, PLANET v2.0 was tested without re-training and fine-tuning. Figure 3 presents a comparison of the results from Glide SP, ConBAP, GenScore, PLANET v1.0, and PLANET v2.0. More details can be found in Table S4 in the *Supporting Information*. Glide SP is recognized as an efficient tool in virtual screening and is widely applied in drug discovery protocols, which was taken as the baseline. In this test, Glide SP achieved the mean of 0.536 for AUROC (area under receiver operating characteristic curve), while all deep learning methods demonstrated superior performance, with GenScore achieving the best results (mean AUROC = 0.607). However, as a rescoring model in a virtual screening workflow, GenScore requires pre-generated conformations from molecular docking, meaning its performance is inherently dependent on the initial filtering by scoring functions in molecular docking. When it comes to a decoy, molecular docking will not produce a reasonable binding pose, so this decoy can be identified by the model easily. PLANET v2.0 ranked second with a mean AUROC of 0.576, demonstrating improved performance compared to its predecessor (mean AUROC = 0.556). Compared to conventional function Glide SP, PLANET v2.0 produced higher AUROC scores on 8 targets, comparable scores on 3 targets, and lower scores on 4 targets in LIT-PCBA. However, $EF^{1\%}$ of PLANET v2.0 is not better than other models, which may be caused by the function formula of summation. PLANET v2.0 tends to produce a higher predicted value for compounds that have more heavy atoms, although our model has good ranking power among compounds with similar molecular weights. Nevertheless, PLANET v2.0 is a tool which can work with preliminary docking, so that it can be employed to screen ultra-large-scale libraries where conventional docking approaches are computationally prohibitive.

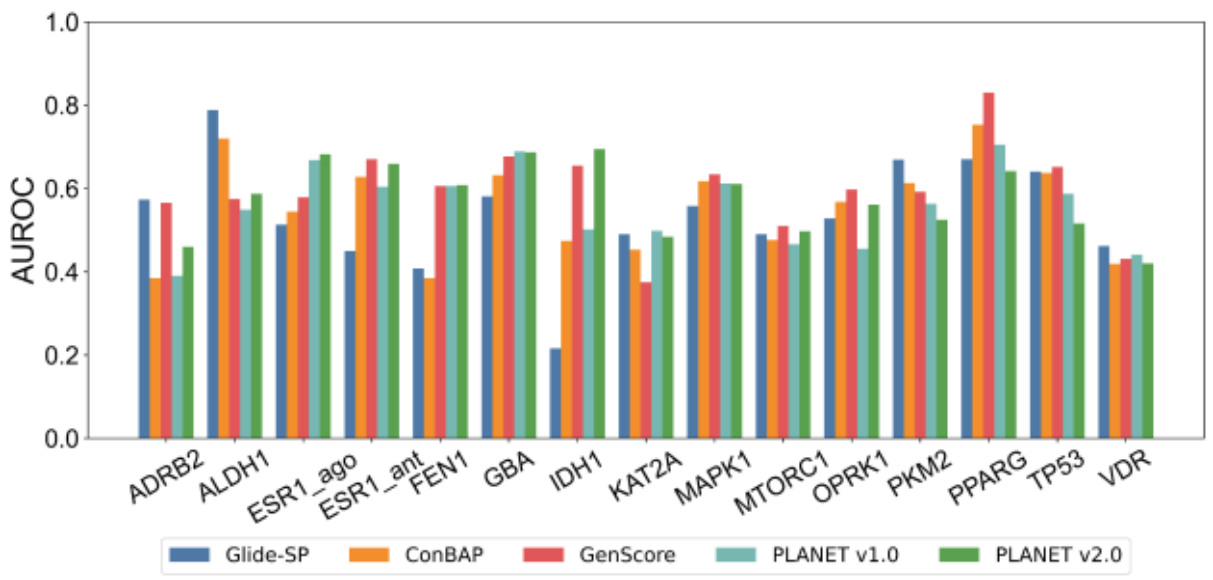

**Figure 3** AUROC comparison of Glide SP and several deep learning methods on 15 targets of LIT-PCBA benchmark.

Although each target in LIT-PCBA contains 170,000 molecules on average, it still falls short in terms of data scale compared to real virtual screening scenarios, failing to fully demonstrate the potential of deep learning models. As a result, we developed a modified LIT-PCBA dataset by using molecules in a commercial dataset as decoys. This commercial compound library contains approximately 19 million compounds, enabling the evaluation of our model's performance in ultra-large-scale virtual screening. Origin actives of 15 targets are still defined as their actives. Results are shown in Figure 4 and Table S5, where the averages of AUROC and $EF^{1\%}$ were 0.463 and 1.115. There is no doubt that this modified dataset is a more difficult task for PLANET v2.0. In order to eliminate the effect of correlation between predicted affinities and molecular sizes, we built up a virtual screening protocol by combining PLANET v2.0 and filters of druglikeness. Lipinski's Rule of Five (Ro5) and ligand efficiency (LE) are both typical filters in drug discovery. When only using Ro5 to filter compounds before scoring, the two metrics both got improved to 0.501 and 3.627, revealing that there was a potential correlation between molecular sizes and predicted values indeed. Then predicted affinities were divided by the numbers of heavy atoms respectively, one can see that AUROC and EF1% further improved to 0.522 and 4.001. The enrichment factor exhibited a moderate decline on some targets, attributable to the tendency of deep learning models to generate prediction values confined to a narrow range for individual targets. The ligand efficiency metric, functioning as a normalized indicator, may further compress score differentials between molecules, thereby blurring the distinction between active compounds and decoys. Although screening power of PLANET v2.0 still has room for improvement, PLANET v2.0 still demonstrates the potential to integrate with traditional virtual screening techniques, and it should be noticed that PLANET v2.0 can finish this task in 10 hours upon the regular computational platform, consisting of a single AMD EPYC9654 CPU and a single NVIDIA GeForce RTX4090 GPU. It takes about 1.5 ms to predict the affinity for one protein-ligand complex, which is extremely faster than conventional tools. Based on these findings, PLANET v2.0 is currently insufficient as a stand-alone screening tool, but it can help narrow chemical space prior to docking-based virtual screening.

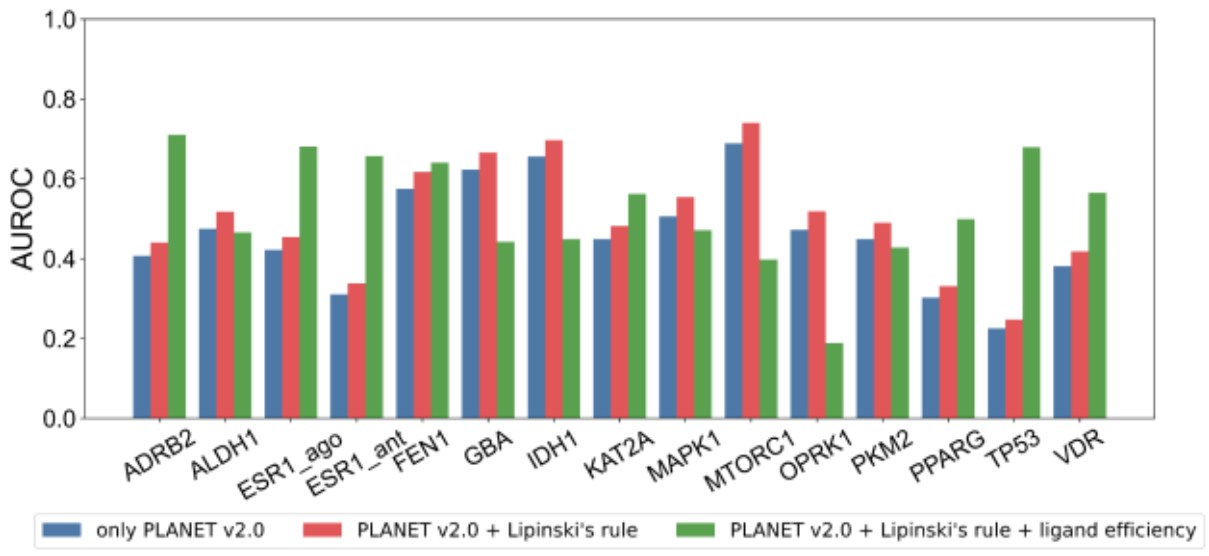

**Figure 4.** AUROC comparison of PLANET v2.0 and combined screening protocols on the modified LIT-PCBA dataset.

### 2.3 Sensitivity to the affinity difference between structural analogs

Except for virtual screening, hit-to-lead or lead optimization are also application scenes for binding affinity prediction models, which requires models to identify the structure-affinity-relation between compounds. Both activity changes and cliffs can exist in two scenarios, so it is a comprehensive challenge based on scoring power and ranking power. There are two widely recognized benchmarks in this field, and they are designed to evaluate the precision of molecular-dynamics-based relative free energy calculations, referred as FEP+ benchmark[71] and Merck FEP benchmark[72] here. In order to facilitate comparison with other methods, Spearman *R* were used as metrics for the two datasets respectively. Results of evaluation on FEP+ benchmark was detailed in Table 3. Among all 8 targets, PLANET v2.0 performed well on 5 targets, where it ranked at the top three of deep learning methods. Our method only performed poorly on one target, but the result only exhibited low correlation, unlike those results from some models which showed no correlation or even negative correlation. In terms of overall correlation, it ranked inferior to the free energy perturbation method, with a Spearman correlation coefficient of 0.64. It is needed to emphasize again that our model cannot directly perceive the difference between binding modes of structural analogs because of the input format, which is a significant disadvantage for this task. Merck FEP benchmark was then utilized to assess our model's sensitivity to structural analogs (Table 4). Differently, PLANET v2.0 performed moderately with an average Spearman *R* of 0.400, close to the widely applied function Autodock4. As one can see, PLANET v2.0 significantly outperforms other methods on some targets, such as c-Met, and Eg5, underscoring the availability of PLANET v2.0 on specific targets. Moreover, although the training sets of most deep learning-based models in the table are derived from PDBbind, their performance across these targets shows no correlation. This observation indicates that that deep learning methods indeed have an effective scope of application.

**Table 3.** Ranking powers of several representative scoring functions across eight targets on the FEP+ benchmark.

| Target | Method | | | | | | | | | |
|---|---|---|---|---|---|---|---|---|---|---|
| | DeepDock | PIGNet | $K_{deep}$ | 3D-GNN | TANKBind | RTMScore | EquiScore | Glide SP | FEP+ | PLANET v2.0 |
| BACE | -0.13 | -0.20 | 0.42 | 0.41 | -0.19 | -0.05 | 0.52 | 0.11 | 0.74 | 0.44 |
| CDK2 | 0.11 | -0.41 | 0.84 | 0.80 | 0.73 | -0.42 | 0.78 | -0.36 | 0.41 | 0.81 |
| JNK1 | 0.62 | 0.70 | 0.45 | -0.51 | 0.41 | 0.66 | 0.62 | 0.24 | 0.90 | 0.64 |
| MCL1 | 0.49 | 0.13 | 0.45 | 0.21 | 0.70 | 0.45 | 0.42 | 0.50 | 0.78 | 0.51 |
| p38 | 0.24 | 0.03 | 0.37 | 0.48 | 0.51 | 0.66 | 0.53 | -0.24 | 0.64 | 0.27 |
| PTP1B | -0.36 | -0.12 | 0.70 | 0.75 | 0.76 | 0.45 | 0.59 | 0.23 | 0.82 | 0.80 |
| Thrombin | -0.07 | 0.91 | 0.75 | 0.55 | 0.80 | 0.81 | 0.64 | 0.49 | 0.62 | 0.70 |
| Tyk2 | -0.13 | 0.69 | 0.52 | 0.71 | 0.49 | 0.59 | 0.36 | 0.79 | 0.87 | 0.41 |
| Mean | 0.096 | 0.216 | 0.562 | 0.425 | 0.526 | 0.394 | 0.558 | 0.220 | 0.722 | 0.572 |

**Table 4.** Ranking powers of several representative scoring functions across eight targets on the Merck FEP benchmark.

| Target | Method | | | | | | | | | | |
|---|---|---|---|---|---|---|---|---|---|---|---|
| | AutoDock4 | Vina | Vinardo | Glide SP | Glide XP | MM-GBSA | $\Delta_{Lin_F9}$XGB | GenScore | DeepRLI | ConBAP | PLANET v2.0 |
| CDK8 | 0.629 | 0.849 | 0.782 | 0.345 | 0.617 | 0.649 | 0.826 | 0.675 | 0.513 | 0.563 | 0.187 |
| c-Met | 0.324 | -0.257 | -0.359 | 0.378 | 0.165 | 0.499 | 0.077 | 0.677 | 0.745 | 0.262 | 0.736 |
| Eg5 | -0.397 | -0.520 | -0.475 | -0.111 | 0.017 | -0.002 | -0.099 | 0.275 | -0.024 | 0.524 | 0.537 |
| HIF-2α | 0.376 | 0.493 | 0.371 | 0.445 | 0.410 | 0.282 | 0.480 | 0.437 | 0.459 | 0.250 | -0.046 |
| PFKFB3 | 0.530 | 0.546 | 0.515 | 0.480 | 0.513 | 0.554 | 0.603 | 0.571 | 0.577 | 0.398 | 0.339 |
| SHP-2 | 0.609 | 0.569 | 0.490 | 0.542 | 0.490 | 0.585 | 0.640 | 0.338 | 0.639 | 0.231 | 0.658 |
| SYK | 0.544 | 0.519 | 0.379 | -0.006 | 0.124 | 0.108 | 0.103 | 0.144 | 0.441 | 0.447 | 0.527 |
| TNKS2 | 0.558 | 0.538 | 0.305 | 0.316 | 0.582 | 0.158 | 0.458 | 0.578 | 0.331 | 0.049 | 0.265 |
| Mean | 0.397 | 0.342 | 0.251 | 0.299 | 0.365 | 0.354 | 0.386 | 0.462 | 0.460 | 0.340 | 0.400 |

## 2.4 Understanding of protein-ligand interaction

PLANET v2.0 was trained via the multi-objective training strategy. Except for prediction of binding affinity, PLANET v2.0 can predict ligand binding conformation in a way of binary classification and model protein-ligand interaction pairs by Gaussian density functions. These objectives are separate from each other but the following two are indispensable, which help the model extract complex structural features from uncorrelated protein and ligand embeddings. The special working mechanism requires PLANET v2.0 make precise affinity predictions on the premise of correct predicted binding modes. 'Docking power' test of CASF-2016 contains 285 complexes together with 100 conformational decoys derived from molecular docking procedures per target. This test can be used to evaluate the basic functionality of PLANET v2.0 as a scoring function and can also serve as a validation of the method's interpretability.

As PLANET v2.0 is an end-to-end model which predicts affinity based on two-dimensional ligand graph, the final predicted values do not change due to conformational variations. Thus, a statistical potential aggregated by all pairwise probability density functions was taken as the score to rank different conformations. The formula can be seen in equation 10 in *Methods*, which is an addition of all negative log-likelihood values calculated for each protein-ligand pair. The results interpreted that PLANET v2.0 achieved the success rates of 85.2%, 94.7% and 97.2% in the top 1, top 2 and top 3 rankings, respectively (Figure 5A). If the native binding pose was not included into the test set, the top-ranked pose was within RMSD < 2 Å in 78.1% cases and the success rate increased to 93.6% when considering the three top-ranked poses (Figure 5B). Based on these results, PLANET v2.0 ranked the best of all methods in CASF-2016, and was better than DeepDock. It is worth noting that PLANET v2.0 ranked the crystal conformation in the first place in almost all successful cases. Models based on MDN were superior to other methods, showing that Gaussian density functions is suitable for representing protein-ligand interactions. In addition, the top 1 success rate of PLANET v2.0 was lower than other MDN-based methods, which might be caused by decoys used in our training set. As shown in Equation 9, the loss function for decoys can influence the shape of the distance distributions predicted by the MDN. Another possible reason is that the pairwise distance probability distribution was simultaneously optimized by both the L2 loss for affinity prediction and the MDN loss, where complex gradient interactions affected convergence in this task. Nevertheless, it is illustrated in Figures2 that the Spearman $R$ between conformational RMSDs and scores of 285 targets still excelled in those of functions in the CASF-2016 benchmark.

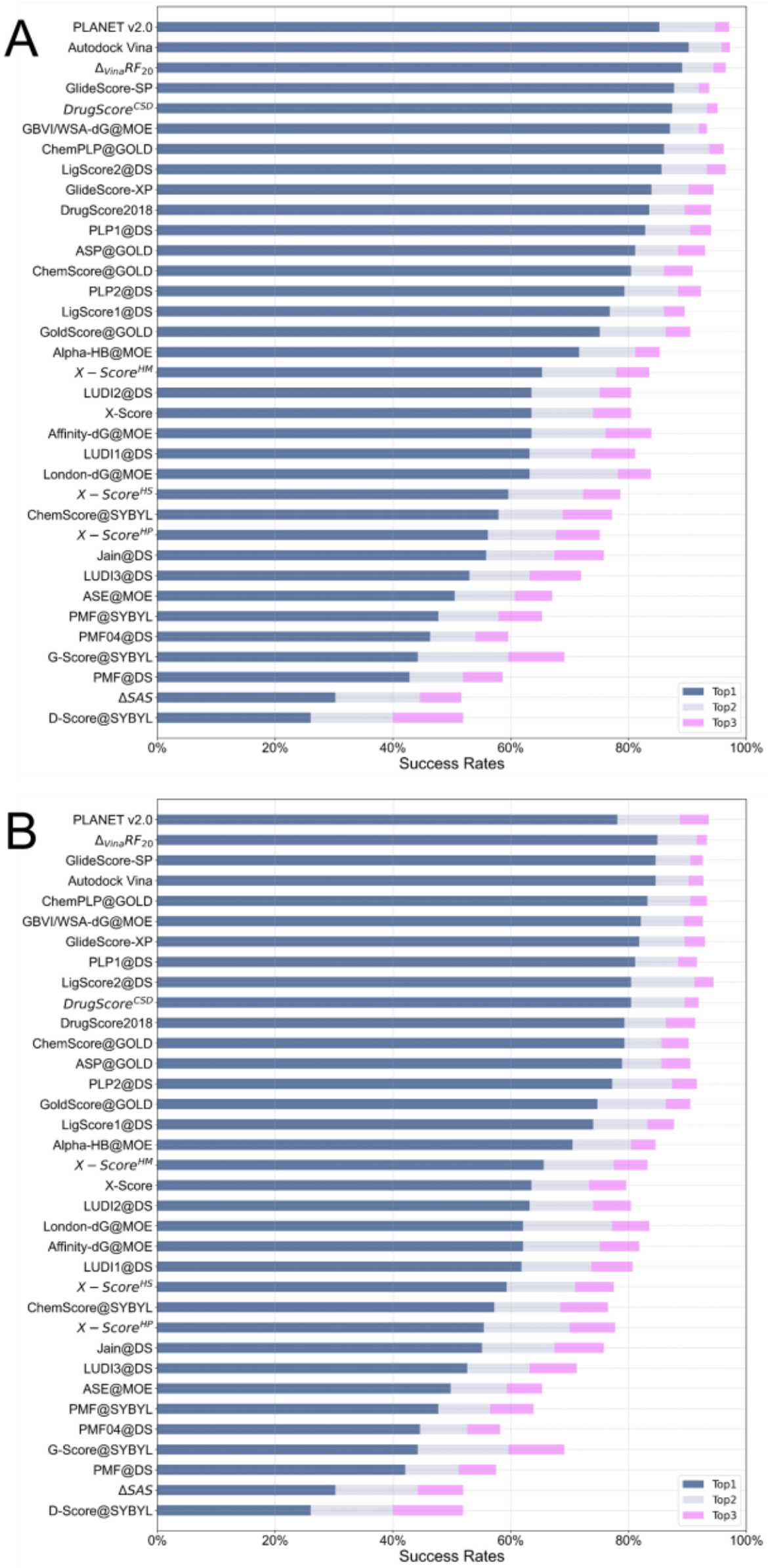

**Figure 5.** Results of docking power test and visualization of ligand binding conformation predicted by PLANET v2.0. (A) Docking power of PLANET v2.0 on the CASF-2016 benchmark, where success rates of including the ligand binding pose (RMSD < 2 Å) among the top one to three candidates are illustrated. (B) Success rates of methods if the native ligand binding pose is not mixed into the decoy set.

The precision of ligand binding conformations predicted by PLANET v2.0 is measured by cosine similarity between the matrix derived from the real structure and the predicted one. Among all of 285 complexes in CASF-2016, PLANET v2.0 produced an average of 0.959, showing that our model successfully built a bridge between two-dimensional graph and three-dimensional conformation and the predictions from PLANET v2.0 were interpretable. Here, some comparisons are presented in Figure 6, which were selected based on atomic count of ligands. It should be mentioned that there are few deep learning models with good docking power also excel in the other three powers. Some tricks had to be employed to address this issue, such as adding correlation values to the training loss or integrating a model with another output module. In this regard, PLANET v2.0 is a competitive model and the first end-to-end framework that balances all four powers.

**Table 5.** Performance of PLANET v2.0 and a baseline model in ligand distance matrix on the CASF-2016 test set

| Model | Metrics | | |
|---|---|---|---|
| | Cosine similiarity | AUROC | AUPRC |
| PLANET v2.0 | 0.959±0.051 | 0.986±0.033 | 0.980±0.060 |
| PLANET v1.0 | 0.925±0.049 | 0.975±0.033 | 0.961±0.061 |
| baseline | 0.519±0.105 | 0.500±0.017 | 0.385±0.131 |

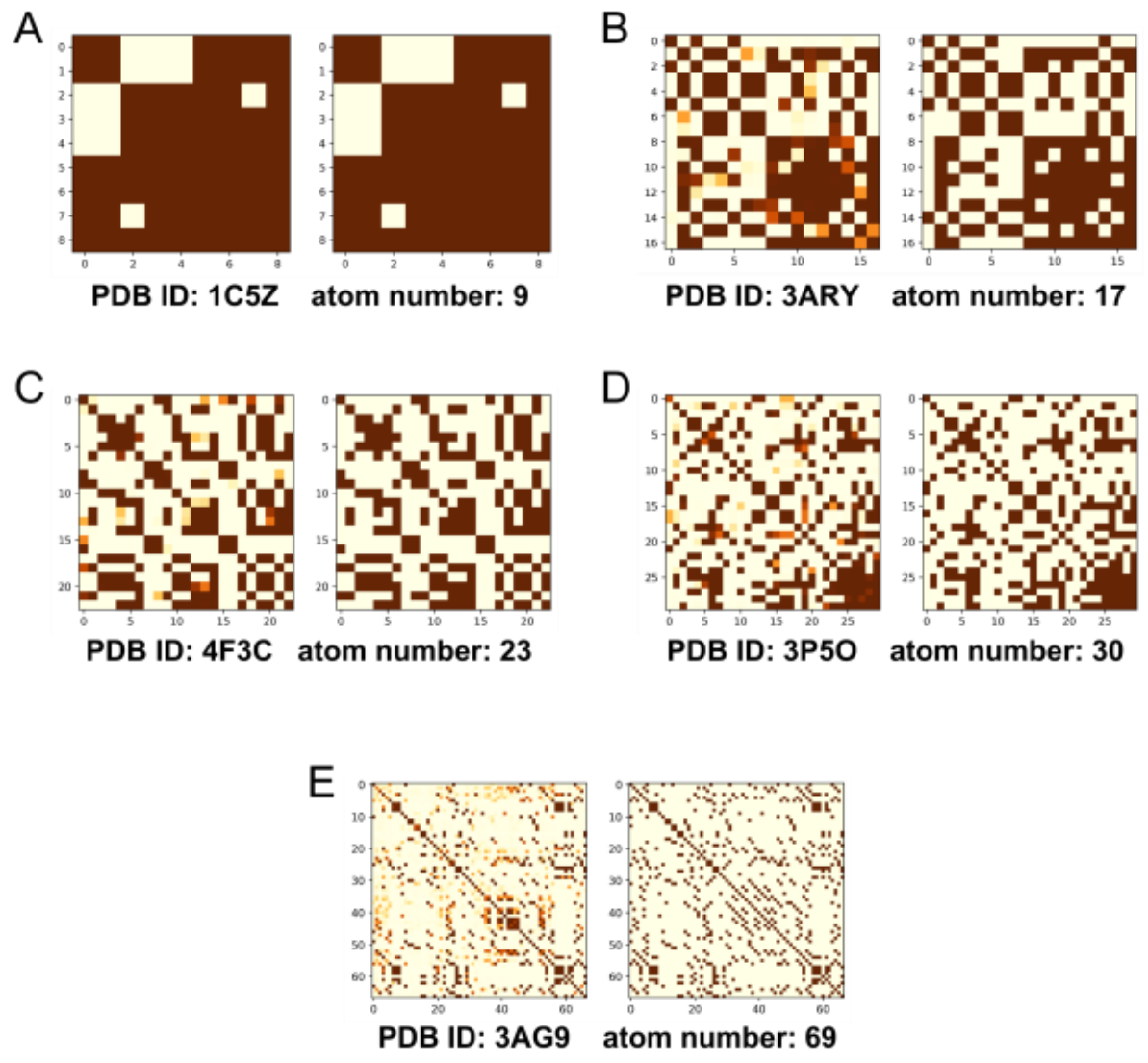

**Figure 6.** Five examples of the predicted ligand binding conformations, whose heavy atom numbers are the minimum (A), first quartile (B), median (C), third quartile (D), and maximum values (E) of all ligand atomic counts, respectively.

Furthermore, we carried out a binding funnel analysis for PLANET v2.0, and the comparison result between PLANET v2.0 and scoring functions in CASF-2016 was shown in Figure7. A red region at the short RMSD range (e.g., RMSD < 5 Å) is expected, which means that there is an obvious funnel on their binding energy surface. As one can see, PLANET v2.0 had good docking 'efficiency' and 'accuracy', because the result of PLANET v2.0 showed a strong correlation within this range. Like X-Score and its variations, PLANET v2.0 also produced a strong correlation at a wide RMSD range, indicating that PLANET v2.0 was hopeful to be applied in a blind docking job.

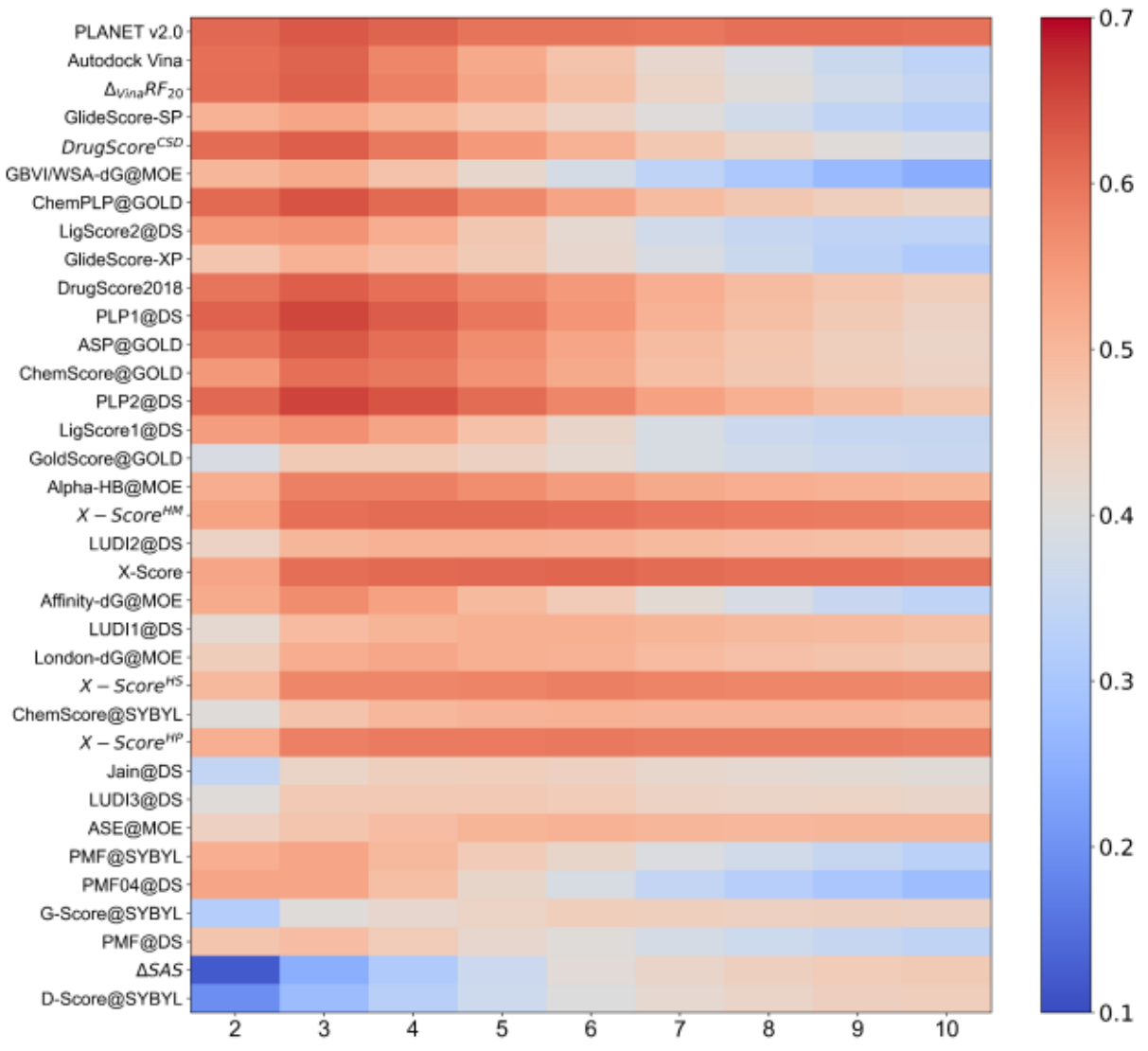

**Figure 7.** Results of the binding funnel analysis for PLANET v2.0. The x-axis indicates the RMSD range (e.g., [0–2 Å], [0–3 Å], and so on) where the Spearman correlation coefficients between RMSDs and scores are computed. For the convenience of making a comparison, all scoring functions are ranked in the same order as in Figure 5A.

**2.5 Impact of modules on performance of affinity prediction**

Typically, there are two approaches of ablation study, including ablating modules and strategies used in development. Take PLANET v1.0 as an example, we demonstrated the importance of training set and multi-objective training to our model. Thus, similar ablation studies are not presented here. Ablation studies of key modules were carried out in many researches, but we worried that the declining performance observed in this way might be caused by a reduction in the number of learnable parameters, where features did not get necessary projection. Thus, we focused on two protein feature extraction modules and protein-ligand communication module in our studies, so that key functions were not missing in the ablated models. The models in Table 5 were as follows: (1) '*Pv2_baseline*': This model was the complete PLANET v2.0; (2) '*Pv2_no_glob_feat*': This model did not have the module of global pocket feature extraction; (3) '*Pv2_no_loc_feat*': This model did not have the module of local residue feature extraction; (4) '*Pv2_no_comm*': This model did not have protein-ligand communication module; (5) '*Pv1_2020*': This model was the original PLANET which we reported before, and it was trained with PDBbind v2020; (6) '*Pv1_2021*': This model was PLANET which was retrained with PDBbind v2021 to facilitate comparison with series of PLANET v2.0. As one can see, scoring power results of all ablated models decreased to coefficients around 0.830, showing an obvious gap toward *Pv2_baseline*. It is indicated that features at different scales were crucial for PLANET v2.0 to establish the relation between two-dimensional representations and three-dimensional features. Compared to our last model, even ablated models of PLANET v2.0, which shared the main architecture of MDN in common, showed improvement in every metrics. This result demonstrated that 'MDN-to-affinity' module was reasonable, and modeling protein-ligand pairwise features with GMMs probably addressed the problem of imbalanced labels in PLANET.

**Table 5.** Scoring power of PLANET v2.0 in several ablation studies.

| Model | Scoring Power on CASF-2016 | | |
|---|---|---|---|
| | MAE | RMSE | Pearson $R$ |
| *Pv2_baseline* | 0.908 | 1.171 | 0.848 |
| *Pv2_no_glob_feat* | 0.953 | 1.229 | 0.829 |
| *Pv2_no_loc_feat* | 0.970 | 1.249 | 0.826 |
| *Pv2_no_comm* | 0.963 | 1.241 | 0.834 |
| *Pv1_2020* | 0.965 | 1.247 | 0.824 |
| *Pv1_2021* | 0.956 | 1.235 | 0.829 |

**2.6 'Scoring power' tested on new data of PDBbind v2024**

PDBbind v2021 was based on experimental structures available from PDB in November 2020, and CASF-2016 was reorganized from PDBbind v2016. As we all know, there has been a significant shift in the research focus of drug discovery in recent years, such as PROTACs and virus-targeted molecules. Therefore, CASF-2016 may have potential biases and cannot reflect the performance of scoring functions in current drug discovery. As a result, new data in PDBbind v2024 was chosen to assess several models in this aspect. In addition to our model, we selected $\Delta_{\text{vina}}RF_{20}$, the top-performing scoring function in the assessment of CASF-2016, as the baseline model.

RF-Score, pafnucy and GIGN were chosen as representative methods for machine learning, CNN, GNN methods, respectively. RTMScore and GenScore were considered as MDN-based models like ours, and GenScore is an updated version of RTMScore, which was trained using the correlation between experimental affinities and predicted values as its loss function, so the result of GenScore was more comparable. Figure 8 showed that scoring power of each model exhibited varying degrees of decline, and PLANET v2.0 remained relatively the best (Pearson $R = 0.696$). From the distribution of predicted values, the predictions of other deep learning models were concentrated within a specific range, while PLANET v2.0 could make predictions beyond this range. As a result, the slope of our regression line is the closest to 1. Within MDN-based methods, the lack of scoring power of RTMScore was expected as it was designed solely for rescoring purposes, but the result of GenScore still showed no significant improvement. This suggested that the modified training protocol might be ineffective, and its apparent performance gained on CASF-2016 could simply stem from the similarity between training and test sets, rather than the methodological advancement. Surprisingly, RF-Score, which was reported in 2007, still demonstrated remarkable performance in this test set. It was indicated that its atomic-level protein-ligand interaction representation showed lower data dependency and retained practical utility compared to complicated deep learning representations. Under current data constraints, excessive innovation in model architecture appears unwarranted, developing more scientifically grounded representations should be the primary direction for advancing scoring functions, instead.

Apparently, there is still much room for PLANET v2.0 to improve its scoring power within this test set, it is probably because the test set is less homologous with the training set than CASF-2016. The visualization of datasets is detailed in Figure S3. There is no significant difference in molecular fingerprints and binding modes between the two datasets, while the number of protein clusters, clustered at 90% identity, increased by approximately 20% (Table S6). The result suggested that difference in proteins may be the primary reason for the performance decline, and subsequent models should be developed with the latest PDBbind dataset.

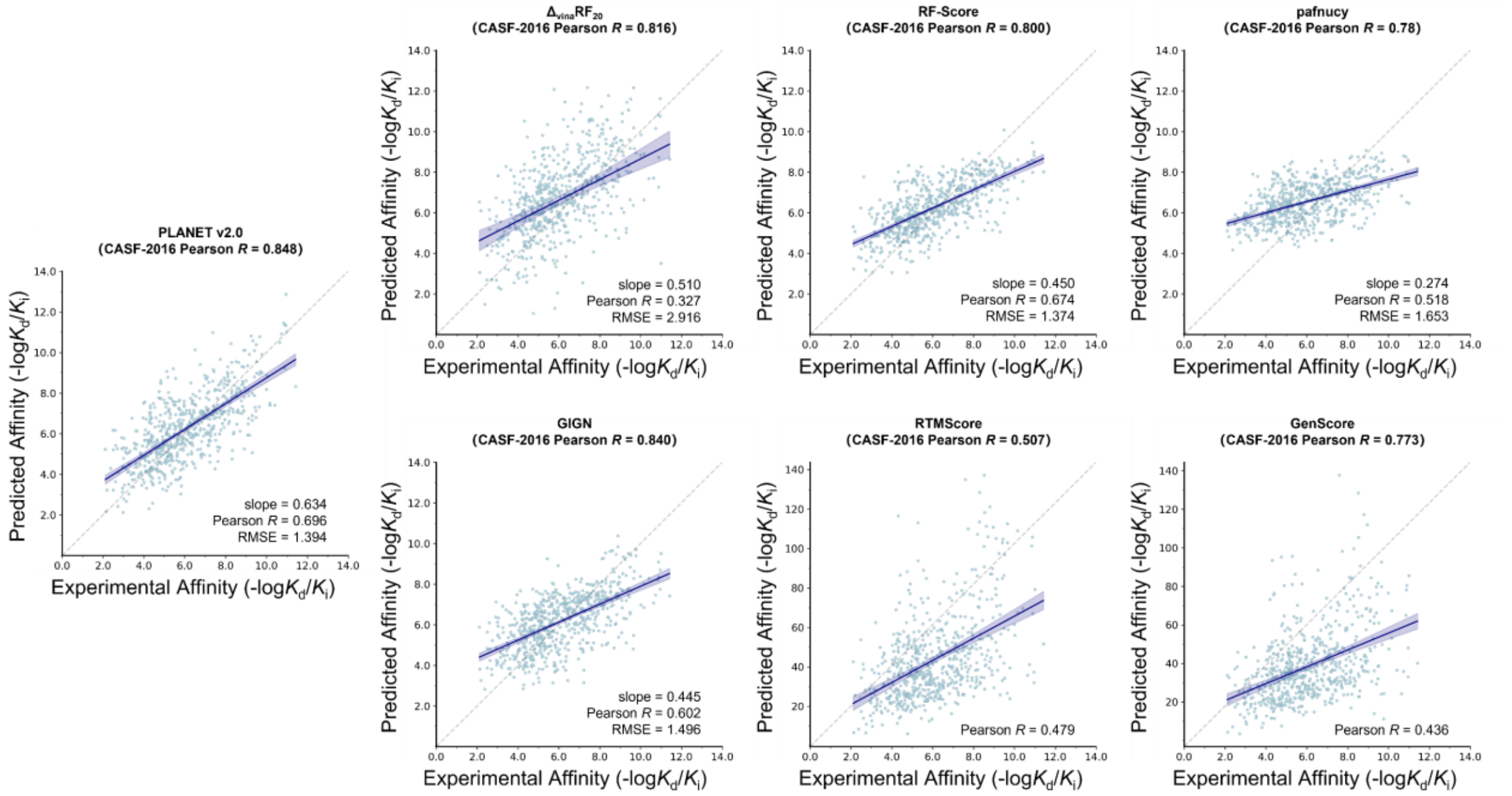

**Figure 8.** Scoring performance of five models on the test set from new data in PDBbind v2024. The solid line is the regression line, where the 95% confidence interval is indicated by the shaded region. The number of samples is different because it is contingent upon the success of feature extraction.

## 3. Conclusion

In this work, we developed an all-round protein-ligand affinity prediction model, called PLANET v2.0. As an updated version of PLANET v1.0, the new model takes the advantages of multi-objective training. It directly extracts input features from three-dimensional structure of protein binding pocket and ligand structures, and then predicts a series of means, standard deviations, and weighted coefficients for each protein-ligand node pair. These parameters are used to parametrize probability density functions and distance-energy functions in the form of GMMs, which further act as prior knowledge of affinity prediction. PLANET v2.0 systematically solves the problems PLANET v1.0 and DeepDock. PLANET uses binary labels to depict interaction pairs of protein and ligand nodes, however these labels are extremely unbalanced, making the model confused during training period, so the binary task is cancelled in PLANET v2.0 and distributions from MDN are applied. As mentioned in the Introduction section, the architecture of MDN is inspired by DeepDock. We expand the description of density distributions by incorporating predictions of energy functions in addition to density functions, and pairwise energy terms can be obtained through expectation calculations. When generating energy functions, only geometric characteristics of the Gaussian mixture model are considered, where mixture coefficients can be either positive or negative, allowing both favorable and unfavorable interactions to be represented simultaneously. In this way, scoring power deficiency in MDN-based models can be resolved.

The results on the popular CASF-2016 benchmark demonstrate that PLANET v2.0 ranks among the top in terms of scoring power, ranking power, and docking power. Additionally, the precision of binding conformation prediction indicates that PLANET v2.0 exhibits strong interpretability. The screening power of PLANET v2.0 was evaluated on the LIT-PCBA benchmark, showing further improvement compared to Glide SP and PLANET. We also conducted an additional test by constructing an ultra-large virtual screening dataset to validate the performance of PLANET v2.0 in real-world screening scenarios and explored the potential of integrating our model with conventional druglikeness filters. In tests such as lead optimization, PLANET v2.0 also demonstrates potential for application. Although PLANET v2.0's performance degraded on new data of PDBbind v2024 refined set, it remains one of the best deep learning-based affinity prediction models available. In summary, PLANET v2.0 is the first all-round end-to-end protein-ligand affinity prediction model, and we expect it to become a practical tool in drug discovery.

## 4. Methods

### 4.1 Dataset preparation

PDBbind[73] was used as the primary dataset in model development. PDBbind is a large database aiming at collecting experimental affinities of biological complexes with available three-dimensional structures. It has been a major data source for the development of conventional scoring functions and affinity prediction models. The latest version of the PDBbind database, PDBbind v2021, contains 22,920 protein-ligand complexes. These data were divided into training, validation, and internal test sets in our workflow. As the starting point of data partition, the 285 entries from the CASF-2016 benchmark were firstly selected as the internal test set, and were initially removed from the dataset to prevent information leakage between the training and test sets. Furthermore, we removed 'softly overlapping' data from the dataset. As we proposed in our previous study, 'soft overlap' between training and test datasets can lead to overestimation of model performance, which is a common issue and is always ignored in regarding studies. The CD-hit software[74, 75] was firstly utilized to cluster all protein sequences in the 'general set' of PDBbind v2021 with a 0.9 sequence identity threshold. Any protein-ligand complex from the dataset that belongs to the same cluster of any protein in the test set was identified as a potential "soft overlap" entry. Then, the Tanimoto coefficients of ECFP4 fingerprints between the ligands in potential "soft overlap" data and the ligands in the same cluster of the test set was calculated. Ligands with a Tanimoto coefficient greater than 0.9, along with their corresponding data, were considered redundant. The above steps prevent information leakage from the test set, while minimizing the loss of data as much as possible. Consequently, a total of 400 complexes were defined as "soft overlap" data and removed from our dataset. The remaining data were shuffled and split into training and validation sets at a ratio of 4:1 randomly. Notably, the protein-ligand complex structures provided by PDBbind v.2021 have already undergone structural preparation, including completing missing residues, optimizing energy of the fixed structures, calculating protonation states, and adding hydrogen atoms for proteins, and correcting structures and physicochemical properties for ligands. Therefore, no additional structural processing was conducted during the development of PLANET v2.0. The relevant processing workflows will be reported in our subsequent publications.

In order to enable the model to distinguish between active and inactive ligands, molecules from ChEMBL 31 were selected for each protein-ligand complex and used in training. In our previous work, we compared the physicochemical properties to avoid making the distinction between ligands too simplistic and applied molecular docking to reduce the presence of false-negative molecules. However, due to the computational cost of molecular docking, it is hard to select enough inactive molecules for each complex pair. Therefore, in the data augmentation process of PLANET v2.0 (See Figure S1 in the *Supporting Information*), we only considered drug-likeness and molecular similarity. Each selected molecule was required to meet at least three of the rules and have an ECFP4 Tanimoto similarity of less than 0.2 with the corresponding active ligand. Through these methods rather than molecular docking, approximately one million non-binder decoys were selected as a result, which is five times of the amount used in our previous work. All decoys were preprocessed using the Epik module[76], and the protonation state was set at pH 6.5~7.5.

**4.2 Architecture**

PLANET v2.0 accepts a three-dimensional structure of the protein and a two-dimensional graph of a given molecule as input and then predicts the affinity between them. Note that the model does not use the whole protein structure instead of the active site. The active site, or the pocket, is typically defined by the geometric center of a known bound ligand in PLANET v2.0. If no reference ligand is available, the pocket can be determined manually by coordinates. It is a challenge that affinity prediction models do not directly utilizes the spatial structure of the complex as input, which requires the model to make predictions based on an understanding of protein-ligand interactions, and that is the strength of PLANET. As an upgraded version, PLANET v2.0 has a deeper understanding of binding modes and improves both the modeling and training process of protein-ligand interactions. Both models decompose the protein-ligand interaction into a set of interactions between protein residues and ligand atoms. However, PLANET uses 'hard' labels from the binary classification model to define the relation between a residue and an atom based on distance, which has two main issues. Firstly, it is unreasonable to use the same description method for interactions of different types. Secondly, the number of interaction pairs that satisfy the distance threshold is much smaller than the number of non-interaction pairs. The imbalance in label distribution can complicate the training of the binary classification task. In our assumption, protein-ligand interaction is composed of all residue-atom interaction pairs, each contributing to the binding free energy. In PLANET v2.0, interaction pairs are characterized using an approach like DeepDock. The distributions of probability and energy for each pair are represented through Gaussian mixture models, as the real output of our model. The expected energy contribution of each pair can be obtained without any learnable parameters, ultimately helping to assess the affinity.

The overall architecture of PLANET v2.0 is illustrated in Figure 1, it comprises three modules: graph embedding, protein-ligand communication and protein-ligand interaction prediction.

**4.2.1 Graph embedding and representation**

The protein and ligand features in our model are respectively encoded through two sets of independent graph neural networks. The ligand is represented as a two-dimensional undirected graph, where each node represents an atom, and an edge represents the covalent bond between atoms. The physicochemical properties of the nodes and edges are computed as descriptors and encoded as the initial features in a one-hot format (See Table S1 in the *Supporting Information*). The ligand feature extraction module is composed of an $L_1$ layer of the graph encoder module in JT-VAE[77]. After L1 update iterations (here, $L_1 = 10$), the features of ligand atoms are updated and obtained (Equation 1).

$$m_{x,y}^{l} = f\left(W_{atom} v_x^{init} + W_{bond} b_{x,y}^{init} + W_{message} \sum_{N(x)-1} \widehat{m}_{z,x}^{l-1}\right)$$

$$\widehat{m}_{z,x}^{l-1} = \frac{1}{H}\sum_{h=1}^{H} \alpha_{z,x}^{h,l} W_V^h m_{z,x}^{l-1}$$

$$\alpha_{x,y}^{h,l} = \mathrm{softmax}\left(\frac{W_Q^h m_{x,y}^{l-1} \cdot W_K^h m_{z,x}^{l-1}}{\sqrt{d_K}}\right)$$

$$v_x = f(W_{out}\left(v_x^{init} + \sum_{x \in N(y)} m_{x,y}^{L_1}\right)) \qquad \text{(Equation 1)}$$

where $v_x^{init} \in \mathbb{R}^{36}$ and $b_{x,y}^{init} \in \mathbb{R}^{10}$ denotes the initial embeddings of atoms and bonds, N(·) denotes all neighbors of an atom. Dot (·) denotes the inner product of the matrix, H denotes the number of heads in the multi-head attention mechanism, $f(\cdot)$ denotes a LeakyReLU activation function and W denotes learnable parameters in the linear layer. Notably, $m_{x,y}$ means the message passing from atom x to atom y, which should be distinguished from $m_{y,x}$.

The three-dimensional structure of the protein active site is used as input to our model, with the structure determined by the mass centroid of the ligand molecule in the crystal. Any natural amino acids within 12 Å of this centroid are considered pocket residues. Special substitutions and protonation states of some residues are modified to the respective natural residues, while non-natural amino acids are ignored. The spatial positions of residues are defined by the coordinates of the α-carbon and the position of side chain. The position of residue side chain is determined by the geometric center of a subset of atoms on the side chain, which is call functional center in this work, the definitions are provided in Table S3 in the Supplementary Information. The initial feature of residue consists of the corresponding vectors from the BLOSUM62 matrix[78] and the distance between two above coordinates and the geometric center of the ligand. Each pair of distances is mapped to a 15-dimensional vector through a radial basis function (RBF), resulting in a total of 65 dimensions for the initial residue features ($v_i^{init} \in \mathbb{R}^{65}$). The protein pocket is represented as an implicit graph, where each node denotes a pocket residue. All residue features, along with the distances between Cα atoms and the functional centers of pocket residues, are used as input to a DAN (Distance Attention Network) to compute translationally, rotationally, and reflectionally equivariant weights (Equation 2). Then, these weights are in conjunction with weights from the multi-head attention mechanism, the features for all residues are updated for the first time after $L_2$ ($L_2$ = 5) iterations (Equation 3).

$$v_{i,j}^{dist} = RBF(d_{i,j}), j \in N(i)$$
$$w_{i,j}^{dist} = \phi_w([W_{resi} v_i^{init}, W_{resi} v_j^{init}, v_{i,j}^{ca}, v_{i,j}^{sc}])$$
$$v_i^0 = v_i^{init} + \sum_j w_{i,j}^{dist} v_j^{init} \quad \text{(Equation 2)}$$

$$\hat{\alpha}_{i,j}^{h,l} = \text{softmax}\left(\frac{W_Q^h v_i^{l-1} \cdot W_K^h v_j^{l-1}}{\sqrt{K_d}}\right)$$
$$\alpha_{i,j}^{h,l} = \text{normalize}(\hat{\alpha}_{i,j}^{h,l} \cdot w_{i,j}^{dist})$$
$$v_i^l = f(W_{dist} \times \frac{1}{H} \sum_{h=1}^{H} \alpha_{i,j}^{h,l} W_V^h v_j^{l-1})$$
$$v_i = \phi_d([W_{resi} v_i^{init}, v_i^{L_2}]) \quad \text{(Equation 3)}$$

where [·] denotes concatenation of vectors, and $d_i^{ca}$ and $d_i^{sc}$ denotes distances from α-carbon and the position of side chain to pocket center, respectively. $RBF(\cdot)$ denotes a function whose formula is $RBF(d_i) = e^{-\frac{(d_i-C)^2}{2\sigma^2}}, C = \{2,3,\ldots,15\}$. $\phi_e$ denotes two consecutive linear layers with one LeakyReLU activation function and one Sigmoid activation function while $\phi_e$ consists of two linear layers and two LeakyReLU functions. Then, the protein pocket is represented as an explicit graph, each node denotes a residue and each edge exists between a residue pair whose distance between them is less than 6 Å. All nodes are updated for the second time through the Equivariant Graph Convolutional Layers (EGCLs) for $L_3$ ($L_3 = 5$) iterations (Equation 4). EGCL used in the model is modified from the Equivariant Graph Neural Network (EGNN) proposed by Satorras *et al.*[79], which is characterized by translationally, rotationally, and reflectionally equivariant graph neural network computations with high computational efficiency. It is worth noting that the corresponding layers in DAN and EGCL do not share the same parameters for better performance in training.

$$
\begin{aligned}
m_{i,j}^{l} &= \phi_e([v_i^{l-1}, v_j^{l-1}, \parallel x_i - x_j \parallel^2]) \\
e_{i,j} &= \phi_{inf} m_{i,j} \\
m_i^{l} &= \sum_{j \neq i} e_{i,j} m_{i,j}^{l} \\
r_i^{l} &= \phi_h([r_i^{l}, m_i^{l}])
\end{aligned}
\qquad \text{(Equation 4)}
$$

where $\phi_e$ denotes two linear layers with two LeakyReLU functions, $\phi_h$ denotes two consecutive linear layers with one LeakyReLU activation function and $\phi_{inf}$ denotes one linear layer with one sigmoid activation function.

**4.2.2 Protein-Ligand communication module**

To facilitate message exchange between protein and ligand features, we have modified the protein-ligand feature communication module of PLANET v1.0. In this module, the relation between pocket residues and ligand atoms is derived through a cross-attention mechanism, and the computed weights are used for updating the features of both entities. the features of ligand atoms and protein residues are updated and obtained after two iterations (Equation 5).

$$\alpha_{x,i} = \text{softmax}\left(\frac{W_{qk}^{lig} v_x \times W_{qk}^{pro} v_i}{\sqrt{K_d}}\right)$$
$$v_x^{update} = W_{att}^{lig} (\frac{1}{N_{atom}} \sum_{x=1}^{N_{atom}} \alpha_{x,i} W_V^{lig} v_i + v_x)$$
$$v_i^{update} = W_{att}^{lig} (\frac{1}{N_{resi}} \sum_{i=1}^{N_{resi}} \alpha_{i,x} W_V^{pro} v_x + v_i) \qquad \text{(Equation 5)}$$

Where $N_{atom}$ and $N_{resi}$ denotes the number of atoms in the ligand and that of residues in the pocket, respectively.

**4.2.3 Protein-Ligand interaction prediction module**

The final protein and ligand features are input into MDN to model the protein-ligand interaction. After concatenating the features of each atom and each residue, several linear layers are applied for feature projection and outputting the means, standard deviations, and mixing coefficients of the probability density distribution and energy distribution, respectively (Equation 6). In the period of training, different number of Gaussian functions is determined. Finally, PLANET v2.0 utilized 10 Gaussians.

$$v_{x,i}^p = [W_{lig}^p v_x^{update}, W_{pro}^p v_i^{update}]$$
$$\rho_{x,i}^p = \text{softmax}(W_\rho^p v_{x,i}^p)$$
$$\mu_{x,i}^p = \text{ELU}(W_\mu^p v_{x,i}^p) + 1$$
$$\sigma_{x,i}^p = \text{ELU}(W_\sigma^p v_{x,i}^p) + 1.1$$
$$v_{x,i}^e = [W_{lig}^e v_x^{update}, W_{pro}^e v_i^{update}]$$
$$\rho_{x,i}^e = \tanh(W_\rho^e v_{x,i}^e)$$
$$\mu_{x,i}^e = \text{ELU}(W_\mu^e v_{x,i}^e) + 1$$
$$\sigma_{x,i}^e = \text{ELU}(W_\sigma^e v_{x,i}^e) + 1.1 \qquad \text{(Equation 6)}$$

where tanh activation function was used to project the output to (-1, 1), so that both favorable and unfavorable interactions can be modelled. These $\rho^e$ coefficients are not required to sum to 1, because the geometric feature of Gaussian model rather than its probabilistic feature was exploited here.

According to the principle of physical chemistry, the negative logarithm of the protein-ligand complex dissociation constant is linearly related to the binding free energy.

Based on the predicted Gaussian parameters of MDN, the probability density and energy distributions for each interaction pair are represented. Among protein-ligand interactions, the majority part consists of short-range interactions, such as hydrophobic interactions, while long-range interactions, such as ionic bonds, are less frequent. Therefore, our model only considers the energy contribution from short-range interactions when predicting binding affinity. The threshold of interaction distance is finally determined to be 3 Å. Based on the probability distribution and energy distribution of each interaction distance, the short-range free energy contribution $\hat{e}$ of interaction pair is computed in the form of a mathematical expectation, which is also call 'MDN-to-affinity' module in this article (Equation 7). From another perspective, the probability density function of distance works as the partition function of a complex system at the scale of interaction pairs. The function establishes a connection between microscopic distance states and macroscopic thermodynamic properties, thus the computed energy terms implicitly reflect the conformational dynamics of the complex.

$$\hat{e} = \int_0^3 F(x)E(x)dx \qquad \text{(Equation 7)}$$

It is assumed that the total sum of all energy contributions equals the protein-ligand binding free energy when neglecting some long-range interactions, so all partial interaction energies are summed as affinity prediction (Equation 8), which is optimized using an L2 loss function. It is worth noting that the small distance threshold was determined not only to exclude long-range interactions with minimal contributions but also due to the limited accuracy of the predicted probability density distributions.

$$\widehat{pK_d} = \sum_{x,i}^{N_{atom},N_{resi}} \hat{e} = \sum_{x,i}^{N_{atom},N_{resi}} \int_0^3 F(d_{x,i})E(d_{x,i})dd_{x,i} \qquad \text{(Equation 8)}$$

### 4.3 Multi-objective training

PLANET v2.0 is trained through the multi-objective strategy, where these objectives are interrelated and mutually complementary. PLANET v2.0 has three main training objectives, including prediction of protein-ligand binding affinity, derivation of the ligand distance matrix and prediction of atom-residue pair's distance probability density distribution. PLANET v2.0 takes the binding affinity as an only one output and is applied in various scenarios, making it a true "end-to-end" scoring function. As a result, the other two training objectives, aside from affinity prediction, are primarily regarded as outputs during the supervised training process. They enable PLANET v2.0 to extract three-dimensional features from two-dimensional molecular representation and to predict interaction features based on the independent features of the protein and the ligand. It should be mentioned that the task of derivation of the ligand distance matrix is the same as that in PLANET v1.0, related details can be seen in our previous work. In fact, the multi-objective strategy constructs an internal 'molecular docking' process, which reduces the difficulty of affinity prediction.

Also, our model has two auxiliary tasks: fitting the distance distribution of decoys and recovering types of predicting atoms and residues. These tasks are used to determine that decoys do not have effective interactions with protein and to maintain the fundamental features of the nodes if graph neural networks are too deep, respectively. We employed the Adam optimizer with an initial learning rate of 0.0001 to update model parameters, and the learning rate is reduced by a factor of 0.9 every 50,000 batches. The losses of these training objectives were weighted differently (Equation 9):

$$L = w_1 L_{aff} + w_2 L_{lig} + w_3 L_{MDN} + w_4 L_{decoy} + w_{aux} L_{atom} + w_{aux} L_{resi}$$

$$L_{aff} = \frac{1}{N} \sum_{i}^{N} \left(-\log K_d - (-\widehat{\log} K_d)\right)^2$$

$$L_{lig} = \frac{1}{N_{atom}} \sum_{x,y}^{N_{atom}, N_{atom}} -c_{x,y} \times \ln \widehat{c_{x,y}} - \left(1 - c_{x,y}\right) \times \ln\left(1 - \widehat{c_{x,y}}\right)$$

$$L_{MDN} = \frac{1}{N_{atom} \times N_{resi}} \sum_{x,i}^{N_{atom}, N_{resi}} -logP\left(d_{x,i} \middle| v_x, v_i\right)$$

$$= \frac{1}{N_{atom} \times N_{resi}} \sum_{x,i}^{N_{atom}, N_{resi}} -log \rho_{x,i,k} \mathcal{N}(d_{x,i} | \mu_{x,i,k}, \sigma_{x,i,k})$$

$$L_{decoy} = \frac{1}{N_{atom} \times N_{resi}} \sum_{x,i}^{N_{atom}, N_{resi}} P\left(d_{x,i} \le 5 \middle| v_x, v_i\right)$$

$$L_{atom} = \frac{1}{N_{atom}} \sum_{i,a}^{N_{atom}, 10} -c_a \ln \widehat{c_a}$$

$$L_{resi} = \frac{1}{N_{resi}} \sum_{j,r}^{N_{resi}, 20} -c_r \ln \widehat{c_r} \qquad \text{(Equation 9)}$$

where the weight ratios were finally determined as 2: 0.1: 1: 0.2: 0.001: 0.001. The subscript a and r denote categories of multiclass classification in the task of node type recovery.

## 4.4 Model Evaluation

### 4.4.1 Evaluation of scoring power, ranking power and docking power on CASF-2016 benchmark

PLANET v2.0 was evaluated with CASF-2016 benchmark, which is consisted of 285 protein-ligand complexes. These entries are selected from PDBbind refined set based on identity clustering and some other rules, which are usually chosen as a test set for scoring functions. All structures in this set were preprocessed following the same procedure of our training set. There are four evaluation indicators in CASF-2016, including scoring power, ranking power, docking power and screening power. All these tasks are relevant for PLANET v2.0, but we solely carried out the first three to assess the performance of model. The reason that CASF-2016 benchmark was not used to evaluate screening power of our model is the limitation of data quantity.

‘Scoring power’ reflects the ability to predict experimental affinities, and it is the most fundamental and important capability of scoring functions. The deviation between the predicted values of 285 complexes and experimental values was measured by evaluating the linear relationship. The metrics include Pearson correlation coefficient, MAE and RMSE.

‘Ranking power’ refers to whether a scoring function can correctly rank the affinities of different ligands which belong to one target. In this benchmark, there are 57 protein clusters in total. Different from ‘scoring power’, the metric in CASF-2016 is Spearman correlation coefficient, merely calculating the relationship within 5 complexes belonging to the same target. We performed random sampling with replacement on 57 clusters, and the BCa (accelerated bootstrapping) method[80] was used to analyze the confidence intervals across all test sets, and the confidence interval considered in our work is 90%. Results of other methods are derived from references[24, 40, 54].

‘Docking power’ denotes that a scoring function should be able to pick out a correct conformation which is the pose in crystal or poses with an RMSD less than 2.0 Å from the native pose. Since PLANET v2.0 takes the 2D graph as ligand representation, conformational changes in three-dimensional space do not influence the outputs. This means that PLANET v2.0 cannot directly take this test. However, our approach can output distance probability distributions for atom-residue pairs in addition to affinity. Although this output is only used in the multi-objective training, it still has practical significance. The distribution represents the most likely binding distance for a given atom-residue pair when binding. It is assumed that every pair in crystal is located at the optimal distance, so the closer a conformation is to the crystal structure, the more interaction pairs at optimal distances it should has. Thus, a statistical score can be defined easily (Equation 9):

$$U = \sum_{i,j}^{N_{atom},N_{resi}} -logP\left(d_{i,j} \middle| v_i, v_j\right) \qquad \text{(Equation 10)}$$

The proportion of successful cases where the model successfully identifies the correct pose out of the 285 complexes is calculated, called the docking success rate. Three success rates are analyzed: top 1, top 2, and top 3, indicating the capability to rank the correct conformation in the top positions at different levels.

**4.4.2 Evaluation of screening power on LIT-PCBA benchmark**

One of the advantages of ML-based scoring functions is less computational cost, so virtual screening on larger libraries can come into reality. CASF-2016 is insufficient to reflect the true performance of PLANET v2.0 in large-scale virtual screening. Also, decoys in 'screening power' benchmark come from a cross-docking manner, which are not consistent with actual virtual screening scenarios. LIT-PCBA is an unbiased dataset reported in 2020 and is suitable for tests of machine learning or deep learning methods. It contains 15 targets with actives and decoys in a ratio of 1:1000 and 415,225 experimentally validated compounds. In order to evaluate the performance of ultra-large-scale virtual screening, a commercial compound library containing roughly 19 million compounds from TargetMol Chemicals Inc. was chosen as a decoy dataset. These compounds were utilized as decoys in all of 15 targets in LIT-PCBA while active compounds of each target are the same. Protein structures are the highest resolution ones of all structures provided by LIT-PCBA, both proteins and ligands are prepared in the same way of training set. Area under receiver operating characteristic curve and enrichment factors are used as evaluation metrics for 'screening power'. The test results of Glide SP come from our previous work.

**4.4.3 Evaluation on FEP+ and Merck FEP benchmark**

One potential application of affinity prediction models is lead compound optimization, which requires the model to distinguish the activities of chemical analogs. Here, we chose two datasets to evaluate our model's sensitivity to analogs. Spearman correlation coefficient was calculated per target and the average was calculated to facilitate comparison with other methods.

Wang *et al.* proposed a dataset to evaluate their free energy prediction method, named FEP+, in 2015, which includes a broad range of ligands and targets. This dataset has since been widely used in the following years and it is called FEP+ benchmark in our work. The benchmark includes 8 targets and 199 ligands, ligands within every target are structurally similar and their affinities are validated by experiments. The results of FEP+ and other deep- learning-based methods are sourced from Cao *et al.* 's work[42].

A further study was conducted by researchers from Merck KGaA, Darmstadt,

Germany to assess the accuracy of free energy calculations, which includes 8 targets and 264 ligands. Structures of proteins used in our test are determined by the PDB ids provided in this article. The results of other methods are derived from Lin *et al.* 's article[47].

**4.4.4 Evaluation of scoring power on PDBbind v2024**

Compared to PDBbind v2021, which was chosen as our main dataset, 4,483 protein-ligand complexes from studies in 2021-2023 are collected in PDBbind v2024, which can demonstrate performance of scoring functions in current drug design scenarios. The intersection between new data and PDBbind v2024 refined set was selected as an external benchmark for evaluating the real performance of models, containing a total of 655 complexes. The same preprocessing was carried out to remove 'soft overlap' data in this benchmark, in order to eliminate similarity with the training set. The parameters of other models were derived from their articles. The procedures to visualize datasets can be seen in *Supporting Information*.